\documentclass[runningheads]{llncs}
\usepackage{graphicx}

\usepackage{tikz}
\usepackage{comment} 
\usepackage{amsmath,amssymb} 
\usepackage{color}
\usepackage{xspace}
\usepackage{booktabs}
\usepackage{pifont}
\newcommand{\cmark}{\ding{51}}%
\usepackage{siunitx}
\usepackage{multirow}
\usepackage{tabularx}

\makeatletter
\DeclareRobustCommand\onedot{\futurelet\@let@token\@onedot}
\def\@onedot{\ifx\@let@token.\else.\null\fi\xspace}

\def\etal{\emph{et al}\onedot}
\makeatother

\begin{document}
\pagestyle{headings}
\mainmatter

\title{3D Room Layout Estimation \\ Beyond the Manhattan World Assumption}

\titlerunning{3D Room Layout Estimation Beyond the Manhattan World Assumption}
%
\author{Dongho Choi\inst{1,2}}
\authorrunning{Dongho Choi} 
%
\institute{MINDsLab, Inc., Gyeonggi-do, Republic of Korea \and
Seoul National University, Seoul, Republic of Korea\\
\email{dongho.choi@snu.ac.kr}}

\maketitle

\begin{abstract}
Predicting 3D room layout from single image is a challenging task with many applications. In this paper, we propose a new training and post-processing method for 3D room layout estimation, built on a recent state-of-the-art 3D room layout estimation model. Experimental results show our method outperforms state-of-the-art approaches by a large margin in predicting visible room layout. Our method has obtained the $3^{rd}$ place in 2020 Holistic Scene Structures for 3D Vision Workshop.

\end{abstract}

\section{Introduction}
Last decade saw growing attention for recovering 3D room layout from a single image, which would benefit many applications with architectonics, design, virtual and augmented reality. 3D room layout can be viewed as a composition of orientation, corner position, and wall boundaries.

Various works \cite{delage2006dynamic,hedau2009recovering,lee2017roomnet,zhang2014panocontext} have been developed for room layout estimation.
Recent methods train deep neural networks to detect room corners and boundaries.
Specifically, Xu \etal \cite{xu2017pano2cad} estimated
the room layout and the pose of objects by detecting surface normal orientations.
Yang and Zhang \cite{yang2016efficient} predicted the depth from single image to infer 3D room model.
Yang \etal \cite{yang2019dula} performed semantic segmentation of floor plan segmentation.
Although these methods can predict objects in the room, they lack to represent sharp edges and flat wall planes.

\begin{figure}[t]
\centering
\begin{tikzpicture}[scale=0.45]
\fill[fill={rgb,255:red,255;green,249;blue,229}] (6.53, 8.19) -- (11, 8.7) -- (11, 0) -- (3.95, 4.77) -- cycle;
\draw[black, very thick] (3.95, 4.77) -- (0, 4.77) -- (0, 0) -- (8.77, 0) -- (8.77, 8.19) -- (6.53, 8.19);
\draw[black, very thick, fill=red] (1.16, 1.1) circle [radius=0.5];
\node[] at (3.2, 1.1) {Camera};
\fill[fill={rgb,255:red,255; green,220; blue,114}] (3.95, 4.77) rectangle (6.53, 8.19);
\draw[cyan, very thick, dashed] (1.16, 1.1) -- (6.53, 8.19);
\node[] at (5.29, 6.48) {\huge \textbf{?}};

\draw[black, very thick] (16.2, 8) -- +(-4.4, 0) -- +(-4.4, -1.75) -- +(-3.9, -1.75) -- +(-3.9, -1) -- +(-3.5, -1) -- +(-3.5, -2) -- +(-3.2, -2) -- +(-3.2, -3.4) -- +(-4.3, -3.4);
\draw[cyan, very thick, dashed] (13, 4.6) -- ++ (2.58, 3.4);

\draw[black, very thick] (16.2, 4) -- +(-4.4, 0) -- +(-4.4, -1.75) -- +(-3.2, -1.75) -- +(-3.2, -3.4) -- +(-4.3, -3.4);
\draw[cyan, very thick, dashed] (13, 0.6) -- ++ (2.58, 3.4);

\draw[black, very thick] (22.3, 8) -- +(-4.4, 0) -- +(-4.4, -3.4) -- +(-5.6, -3.4);
\draw[cyan, very thick, dashed] (17.9, 4.6) -- ++ (2.58, 3.4);

\draw[black, very thick] (22.3, 4) -- +(-3, 0) -- +(-4.4, -3.4) -- +(-5.6, -3.4);
\draw[cyan, very thick, dashed] (17.9, 0.6) -- ++ (2.58, 3.4);
\end{tikzpicture}

\caption{Illustration of room layout viewed from above. There exist multiple plausible candidates for hidden corners.}
\label{fig:hidden}

\end{figure}
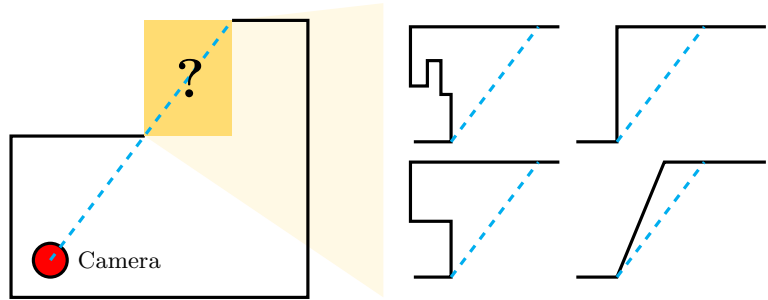

Most common and the best performing method was to predict room boundaries and corners
with Manhattan world assumption \cite{coughlan1999manhattan},
where all wall planes are aligned with three perpendicular axes.
Fernandez-Labrador \etal \cite{fernandez2020corners} and Zou \etal \cite{zou2018layoutnet} predicted the probability map of boundaries and corners as 2D image, while Sun \etal \cite{SunHSC19} predicted as 1D vector. These approaches well predict fine room layouts, but project every objects to wall texture.
However, the Manhattan world assumption limits the potential application
of predicting real world room layout since numerous rooms with non-rectangular shapes exist.
Moreover, these works tried to predict hidden points while hidden points generate numerous layout candidates as in Fig. \ref{fig:hidden}, where all candidates are plausible.

\begin{figure}[t]
	\includegraphics[width=\textwidth]{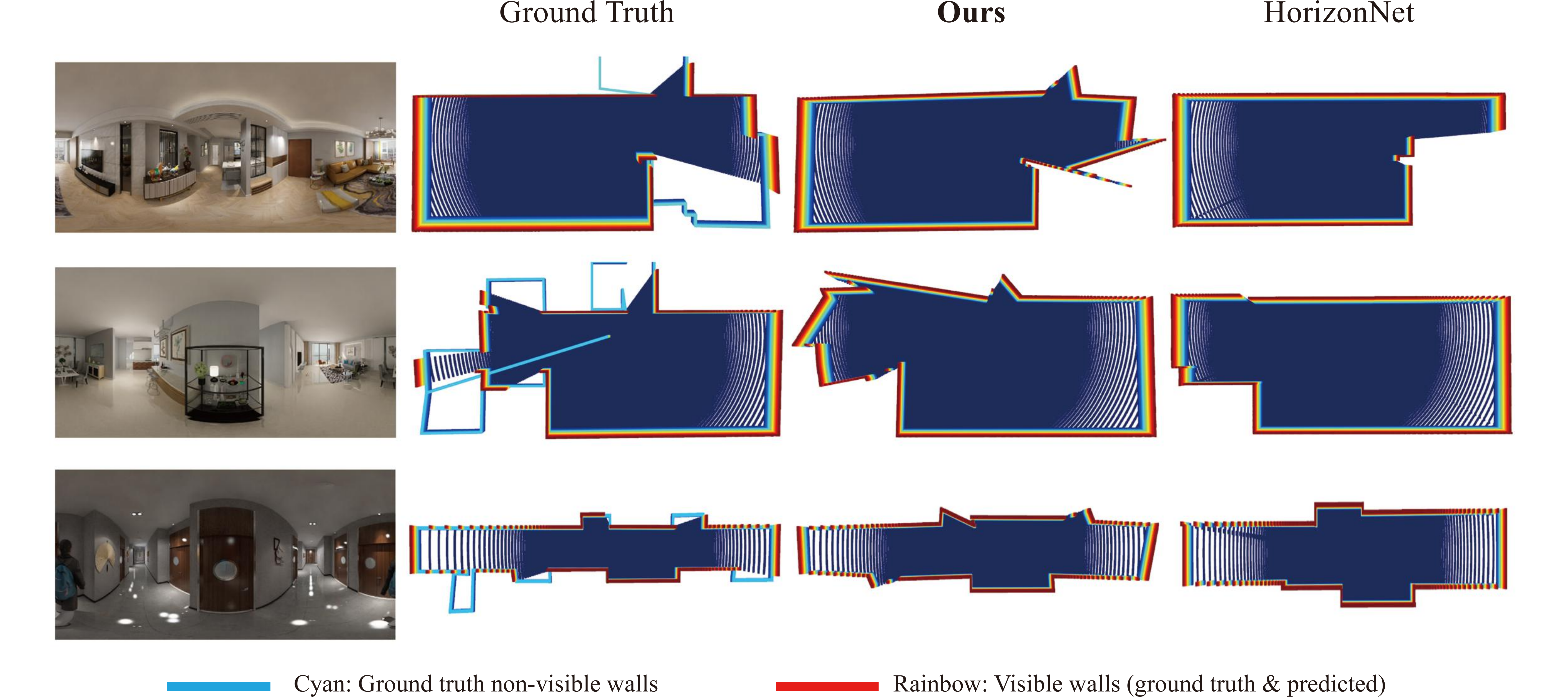}\
	\caption{Illustration of reconstructed 3D layout viewd from above. The leftmost column shows input image, while following columns show ground truth 3D layout, ours and HorizonNet's inference results. Room images are selected from Structured3D \cite{Structured3D}. Best viewed in color.}
	\label{fig:qualitative}
\end{figure}

In this work, we propose a novel training and post-processing method
to predict the visible layout without the Manhattan world assumption.
As shown in Fig. \ref{fig:qualitative},
our method performs better in accuracy than HorizonNet \cite{SunHSC19},
which is the current state-of-the-art method.

\section{Method}
Our model is based on the architecture of HorizonNet \cite{SunHSC19} which predicts 1D layout to recover 3D layout. 
Given a single image \(I\), the network predicts corner probability \(y_p\), ceiling-wall boundary \(y_c\) and floor-wall boundary \(y_f\). 

We improve training and post-processing method while following HorizonNet's network topology and data augmentation method, which was shown to be effective for improving performance by Zou \etal \cite{zou20193d}

\subsection{Loss}
HorizonNet uses L1 loss for boundary prediction while other methods \cite{fernandez2020corners,yang2019dula,zou2018layoutnet} use L2 loss with similar approach. Throughout the community, it has been empirically observed that L1 loss makes the training process unstable due to its constant gradient. Instead, we adopt L2 loss. The linear gradient from the L2 loss helps the network to converge, even when the pixels with small loss dominate.
While L2 loss causes the network to generate blurry boundaries,
we try to predict the corners first and then sharpen the boundaries since boundaries can be accurately calculated from corners.

We give greater weight to the corners during the first half of the training, and swap the weight for boundaries and corners at the last half training time. The overall loss function is as follows:
\begin{equation}
L(y_p, y_c, y_f) = w_1(BCE(y_p, y'_p))  + w_2(L2(y_c, y'_c) + L2(y_f, y'_f))
\end{equation}
where \(y'_p\), \(y'_c\), and  \(y'_f\) represent the ground truth of corner probability, ceiling-wall boundary, and floor-wall boundary, respectively.

\subsection{Post Processing}

\begin{figure}[t]
	\centering
	\includegraphics[width=0.85\textwidth]{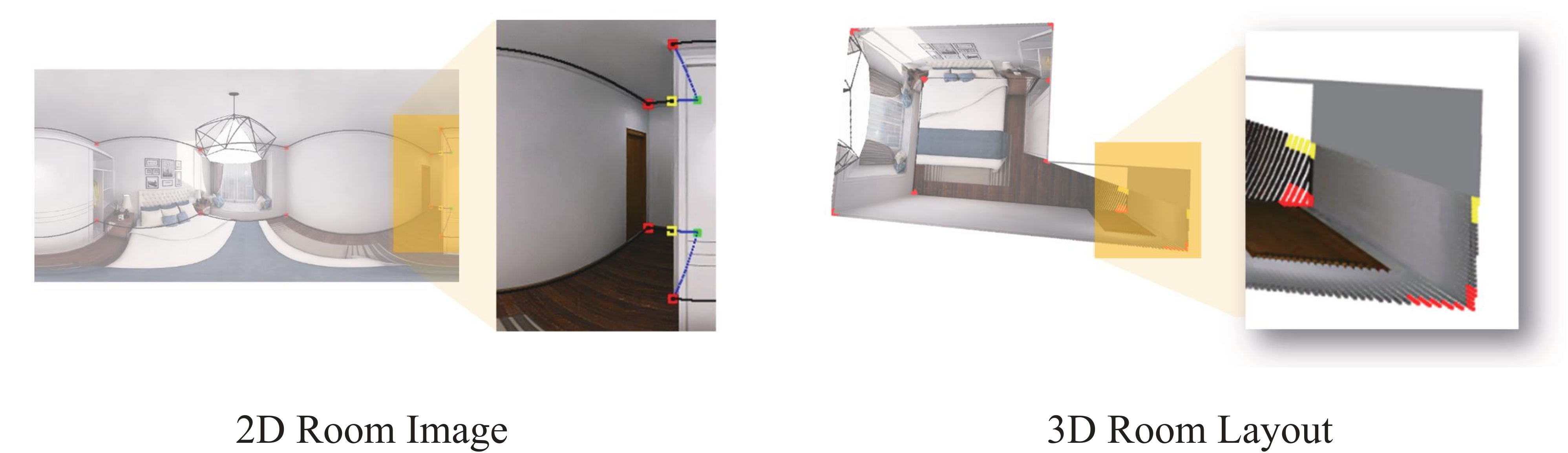}
	\caption{%
		Network output illustrated on 2D, 3D room image.
		Red, green, and yellow depicts the
		continuous visible corners, hidden corners,
		and discontinuous visible corners, respectively.
		The room image is selected from Structured3D \cite{Structured3D}.
		Best viewed in color.}
	\label{fig:post_processing}
\end{figure}

Without the Manhattan layout assumption, hidden corners can not be accurately calculated. We focus on refining visible layouts. In Fig. \ref{fig:post_processing}, visible boundaries are discontinuous near hidden points since walls are not adjacent. Thus, we propose an algorithm to detect discontinuity from predicted boundaries.

We propose two possible approaches for post-processing:
the first directly generates from 2D panoramic image
and the second generates from 3D layout.
We ensemble those candidates and select both the highest and the lowest point for each floor and ceiling boundary at discontinuous points.

\subsubsection{Generating from 2D panoramic image.}
We first detect candidates of discontinuity from raw \(y_c\) and \(y_f\) output. 
In 2D room image, hidden point induces discontinuity in \(y_c\) and \(y_f\) curves. Since the network's output is given by pixels, high slope can be detected as discontinuity. To distinguish between two candidates, we also search for big change in the boundary's slope.

\subsubsection{Generating from 3D layout.}
For 3D layout, distance from the camera to the vertical wall can be accurately calculated when floor and ceiling points are given in 2D panoramic image. As shown in Fig. \ref{fig:post_processing}, hidden point induces jump in distance. We convert 2D panorama image to 3D layout, calculate distance, and find distance discontinuity.

\section{Experiments}
\subsection{Training Details}
\subsubsection{Dataset.} 
We train and evaluate our model using Structured3D \cite{Structured3D} dataset, which consists of more than \(20\)k panoramic images of rooms synthesized with rich details including semantic, albedo, depth, surface normal and input.
We only use single RGB image and corner labels for training. We follow the training, validation, and test set given from the dataset.

\subsubsection{Training Process.}
Our model is implemented with PyTorch \cite{paszke2019pytorch} and tested on a single NVIDIA V100 GPU.
The training process took total 14 GPU days, consisting of 7 days for each training process.
Throughout the training process, Adam optimizer \cite{kingma2014adam} is used with learning rate of \num{3e-4} and \(w_1=3\), \(w_2=1\) for the first half, learning rate \num{1e-4} and \(w_1=1\), \(w_2=3\) for the last.
We train with batch size of 24 and 250 epochs for each half of the training process.
For fair comparison, we apply identical data augmentation strategy with HorizonNet.

\subsection{Quantitative Results}
\begin{table}[t]
\caption{
Quantitative results on room layout estimation from Structured3D test set.
All values are in percentage.
$\dagger$: evaluated with our full post-processing method.
}
\begin{center}
	\begin{tabularx}{\textwidth}{cXcccc}
		\toprule
		\multicolumn{1}{c}{\textbf{Evaluation Layout}}~ & \textbf{Method} & ~\textbf{2D IoU}~ & ~\textbf{3D IoU}~ & ~\textbf{Corner err.}~ & ~\textbf{Pixel err.}~ \\
		\midrule
		\multirow{4}{*}{Non-visible}
		& HorizonNet & 91.72 & 90.17 & 0.860 & -\\
		& HorizonNet$^{\dagger}$ & 91.94 & 90.55 & 0.861 & -\\
		& Ours  & 89.95 & 88.45\ & 0.656 & -\\
		& Ours$^{\dagger}$ & \textbf{93.50} & \textbf{92.20} & \textbf{0.639} & -\\
		\midrule
		\multirow{2}{*}{Visible}
		& HorizonNet$^{\dagger}$ & 92.77 & 91.37 & 0.697 & 2.126 \\
		&Ours$^{\dagger}$  & \textbf{94.31} & \textbf{92.99} & \textbf{0.468} & \textbf{1.340} \\
		\bottomrule
	\end{tabularx}
\end{center}
\label{hidden}
\end{table}

\begin{table}[t]
\caption{
Ablation study on our post-processing method with visible room layout.
The check marks on the 2D, 3D columns represent that the method is evaluated using 2D/3D candidates of our method.
All values are the F scores for each metric.
}
\begin{center}
    \begin{tabular}{lccccc}
    \toprule
    \multicolumn{1}{c}{\textbf{Method}} & & & ~\textbf{Junction}~ & ~\textbf{Wireframe}~ & ~\textbf{Plane}~ \\
    \midrule
     & ~\textit{2D}~ & ~\textit{3D}~ &  &  &  \\
    HorizonNet & \cmark & - & 0.8349 & 0.6655 & 0.9426 \\
    HorizonNet & - & \cmark & 0.8307 & 0.6669 & 0.9430 \\
    HorizonNet & \cmark & \cmark & 0.8382 & 0.6692 & 0.9430 \\
    \midrule
    Ours & \cmark & - & 0.8806 & 0.7380 & \textbf{0.9543} \\
    Ours & - & \cmark & 0.8730 & 0.7378 & \textbf{0.9543} \\
    Ours & \cmark & \cmark & \textbf{0.8834} & \textbf{0.7440} & 0.9542 \\
    \bottomrule
    \end{tabular}
\end{center}
\label{ablation}
\end{table}

Evaluations are based on the following three standard metrics and three F scores: 
\begin{itemize}
	\item IoU: Intersection of Union between our prediction and ground truth.
	\item Corner error: Distance between predicted corners and ground truth corners, normalized by the length of image diagonal.
	\item Pixel error: Pixel-wise semantic(ceiling, wall, floor) error between prediction and ground truth.
	\item Junction, Wireframe: Predicted corner/boundary is considered as correct when prediction and nearest ground truth are within 5, 10, 20 pixels. The final score is an average of three scores.
	\item Plane: Predicted plane is correct when the intersection over union between prediction and nearest ground truth plane is over 0.5.
\end{itemize}
Table \ref{hidden} shows results of layout evaluation on Structured3D dataset.
Our post-processing slightly helps on layout including hidden points even without assuming cuboid layout and predicting hidden points. 

We show ablation study of our post-processing method in Table \ref{ablation}. We obtain large improvement when using candidates from 2D image. Results show that candidates from 3D slightly help to reconstruct room layout.

\subsection{Qualitative Results}

Fig. \ref{fig:qualitative} clearly shows advantages and disadvantages of our method. 
Our method performs better on unusual, non-visible corners but lacks accuracy on rectangular corners since we do not assume perpendicularity between walls.

\section{Conclusion}
In this paper, we proposed a new training and post-processing method for predicting 3D room layout. 
Based on resolving two typical problems of existing works, our method can predict corners from both 2D and 3D room layout and is applicable to other architectures.
Experimental results show the effectiveness of our method. Our method improves state-of-the-art network on predicting visible layout in various quantitative metrics and qualitative results. Although our method shows unsatisfying results on rectangular corners, we believe that future works can improve this by using additional information, multi-image prediction and predicting layout without omnidirectional information.

\clearpage
%
%
\bibliographystyle{splncs04}
\bibliography{egbib}
\end{document}